\documentclass[journal]{IEEEtran}
\ifCLASSINFOpdf
\else
\fi
\usepackage{graphicx}
\usepackage{float}
\usepackage{amssymb}
\usepackage{amsmath}
\usepackage{cite}
\usepackage{multirow}
\usepackage{stfloats}

\begin{document}
\title{Unsupervised CD in satellite image time series by contrastive learning and feature tracking}
%
\author{Yuxing~Chen,~\IEEEmembership{Graduate Student Member,~IEEE} Lorenzo~Bruzzone,~\IEEEmembership{Fellow,~IEEE}
        
\thanks{Y. Chen and L. Bruzzone are with the Department of Information Engineering and Computer Science, University of Trento, 38122 Trento, Italy (e-mail:yuxing.chen@unitn.it;lorenzo.bruzzone@unitn.it).}
\thanks{Corresponding author: L. Bruzzone}}

\maketitle

\begin{abstract}
While unsupervised change detection using contrastive learning has been significantly improved the performance of literature techniques, at the present, it only focuses on the bi-temporal change detection scenario.
Previous state-of-the-art models for image time-series change detection often use features obtained by learning for clustering or training a model from scratch using pseudo labels tailored to each scene. 
However, these approaches fail to exploit the spatial-temporal information of image time-series or generalize to unseen scenarios. 
In this work, we propose a two-stage approach to unsupervised change detection in satellite image time-series using contrastive learning with feature tracking. 
By deriving pseudo labels from pre-trained models and using feature tracking to propagate them among the image time-series, we improve the consistency of our pseudo labels and address the challenges of seasonal changes in long-term remote sensing image time-series. 
We adopt the self-training algorithm with ConvLSTM on the obtained pseudo labels, where we first use supervised contrastive loss and contrsative random walks to further improve the feature correspondence in space-time. Then a fully connected layer is fine-tuned on the pre-trained multi-temporal features for generating the final change maps. Through comprehensive experiments on two datasets, we demonstrate consistent improvements in accuracy on fitting and inference scenarios.
\end{abstract}


\IEEEpeerreviewmaketitle
\section{Introduction}
\IEEEPARstart{D}{etection} of changes in multi-temporal remote sensing (RS) images has been extensively studied in the post decades \cite{liu2019review}.
Early approaches to change detection in bi-temporal RS images include image algebra, image transformation and image classification methods \cite{bovolo2006theoretical}.
These methods have limitations, such as relying on empirical feature extraction algorithms or being sensitive to classification results, which limit their application in change detection.
Image algebra methods directly compare image values, such as in the case of change vector analysis (CVA)-based methods \cite{bovolo2006theoretical, bovolo2011framework, liu2015sequential, zanetti2015rayleigh} that provide spectral change information in terms of magnitude and direction of the spectral change vectors.
On the other hand, image transformation methods map images into the same feature space for comparison.
The most common transformation methods include principal component analysis (PCA) \cite{celik2009unsupervised}, slow feature analysis (SFA) \cite{wu2017post}, and canonical correlation analysis (CCA) \cite{zhang2007remote}.
Supervised image classification methods project image values into different classes at each date and comparedirectly class labels.
This approach, known as post-classification change detection \cite{wu2017post}, is widely used in large-scale land-cover change detection.
In general, image algebra and transformation methods heavily rely on empirical feature extraction algorithms, while post-classification methods are sensitive to the classification results of each image and to error propagation.
These limitations hinder the application of conventional change detection methods.

Deep learning methods have been shown to significantly improve the performance of conventional change detection methods by using deep neural networks \cite{goodfellow2016deep} and stochastic gradient descent \cite{bottou1991stochastic}.
One common approach is direct classification, where models are trained using pre-defined labels and then used to classify change and unchanged pixels.
For example, Rodrigo et al. \cite{daudt2018fully} presented three Unet-based convolution neural network (CNN) architectures for detecting binary changes between pairs of registered RGB images.
In the absence of ground truth, pseudo labels from conventional change detection methods can be used to train models in a self-training paradigm.
Zhou et al. \cite{zhou2020unsupervised} proposed a self-training algorithm based on pseudo labels for change detection, where the pseudo labels are generated by the traditional CVA approach and used to train a new network end-to-end.
The image transformation approach has also been improved using deep learning, where deep neural networks are utilized to extract discriminative features.
Many new techniques have been developed for extracting discriminative features from bi-temporal RS images, such as generative \cite{oussidi2018deep} and discriminative \cite{liu2021self} models.
For the generative model, Luppino et al. \cite{luppino2022code} combined domain-specific affinity matrices and autoencoders (AEs) to align related pixels from multimodal images.
Chen et al. \cite{chen2022self, chen2021self} explored the use of discriminative models in change detection, proposing the use of contrastive learning at pixel-level and patch-level in multi-temporal and multisensor scenarios.
The application of deep learning in post-classification change detection can follow two main directions.
One is to use a deep learning-based segmentation approach to classify the object of interest on bi-temporal images and then compare them.
For example, Nemoto et al. \cite{nemoto2017building} first segmente buildings in an urban area and then compare the building maps at two different times to detect changes.
Another approach is to perform binary change detection and segmentation of both images simultaneously.
Ding et al. \cite{ding2022bi} proposed combining post-classification and direct classification methods using a bi-temporal semantic reasoning network, where the network produces both a change map and two classification maps.
These approaches demonstrate the ability of deep learning in deriving changes from image pairs.

The challenge of detecting changes in remote sensing (RS) images time-series is compounded by the presence of seasonal noise, which can be difficult to distinguish from true changes.
One approach to addressing this challenge is to use graph-based methods \cite{guttler2017graph}, which present detected spatiotemporal phenomena as evolution graphs composed of spatiotemporal entities belonging to the same geographical location in multiple timestamps.
Deep learning methods have also been applied to RS image time-series change detection, using techniques such as recurrent neural networks (RNNs) \cite{lyu2016learning} to extract discriminative features from image sequences.
However, supervised methods often require a large number of labelled training samples, which can be difficult to obtain for long image time-series.
In this context, self-training approaches such as self-supervised and pseudo-label learning have become popular, where networks are trained on a pretext task such as image restoration using 3D CNN \cite{meshkini20223d, meshkini2021unsupervised} and predict the correct order of shuffled image sequences \cite{saha2020change}.
For example, Kalincheva et al. \cite{kalinicheva2020unsupervised} proposed a framework combining a graph model and pseudo labels, which associates changes in consecutive images with different spatial objects using a gated recurrent unit (GRU) AE-based model. 
Meshkini et al. \cite{meshkini20223d} further proposed the use of a pre-trained 3D CNN to extract spatial-temporal information from long satellite image time-series, where they can detect the times and locations of changes in image sequences.
However, pseudo labels often have a high level of noise and do not consider temporal information, and the pre-trained model can not adapt to various changes.

In this work, we propose the use of contrastive learning \cite{khosla2020supervised} and feature tracking \cite{araslanov2021dense} to address these challenges and improve the performance of change detection in RS image time-series.
We leverage contrastive learning methods to get good pre-trained features for pseudo label generation and reduce the overfitting that results in incorrect pseudo labels when considering supervised contrastive learning \cite{khosla2020supervised} and contrastive random walks \cite{jabri2020space}.
Additionally, by incorporating a feature tracking-based pseudo label generation task and a convolutional long short-term memory network (ConvLSTM) \cite{shi2015convolutional}, we are able to extract time-series change maps from image time-series and further train a new model from scratch.
In detail, the pseudo-label generation is based on the pre-trained model using contrastive learning.
The change detection model is trained from pseudo labels by the joint use of Unet \cite{zhang2018road} and ConvLSTM network.
We first extract pseudo labels from change pair time-series and then use them with images to train the proposed network, which outputs change maps relative to the first image in the sequence.
During the training, supervised contrastive loss, contrastive random walk loss and cross-entropy loss are used to optimize the parameters of the feature encoder and the last classifier, respectively.
The supervised contrastive loss is used to mitigate the noise in pseudo labels, while the contrastive random walk loss improves the quality of the consecutive change results.
Finally, we demonstrate the effectiveness of our approach on two data sets.

In this paper, we propose the following main novel contributions:
\begin{itemize}
\item To generate time-related pseudo labels for network training, we propose to use feature tracking to extract reliable change pixels in image sequences that are insensitive to seasonal changes.  
\item To ensure the robustness and consistency of change maps, we propose to use supervised contrastive loss and contrastive random walk loss on change feature learning. These losses encourage the pixels in the same class to have a closer feature representation.
\item To extend the approach to arbitrary long time-series, we jointly use Unet and ConvLSTM as the model architectures. To verify the performance of the proposed approach, we provide a comparison with state-of-the-art methods and an ablation study. Our experiments show that our method obtains competitive results on the datasets. 
\end{itemize}

The remainder of this paper is organized as follows. 
Section II introduces the related works.
In Section III, we introduce our proposed approach, including the network architecture, the supervised contrastive loss, the contrastive random walk loss and the feature tracking-based pseudo label generation. In Section IV, we present the experimental results obtained on two datasets and compare our approach to state-of-the-art methods. In this section, we also include an ablation study in the discussion. Finally, in Section V, we draw conclusions and discuss future work.

\section{Related Works}
\subsection{Self-supervised Learning in Change Detection}
Self-supervised learning is a method of representation learning that does not require human intervention for data annotation, as opposed to supervised learning.
It has been successful in remote sensing image change detection, thanks to its ability to obtain good representative features.
Self-supervised learning has two main streams: generative and discriminative methods \cite{liu2021self}.

Generative models often rely on autoencoders, generative adversarial networks \cite{goodfellow2020generative}, and diffusion models \cite{rombach2022high}.
Denoising autoencoders \cite{riz2016domain}, a classical generative self-supervised learning model, are a type of autoencoder that reconstructs one temporal image from another.
Bergamasco et al. \cite{bergamasco2019unsupervised} proposed the use of a multilayer convolutional denoising autoencoder for unsupervised change detection in multi-temporal Sentinel-1 images.
In addition to autoencoders, generative adversarial networks have also been used for change detection tasks.
For example, Gong et al. \cite{gong2017generative} treated change detection as a generative learning procedure that connects bi-temporal images and generates change maps.
As for diffusion models, Gedara et al. \cite{gedara2022remote} used a pre-trained denoising diffusion probabilistic model to extract feature representations from unlabeled remote sensing images, and then train a lightweight change detection classifier to detect changes from the learned features.

In contrast, discriminative models are mostly based on contrastive learning, which learns a representation that helps distinguishing one object from another.
The objective of contrastive loss is to learn a representation where semantically similar features are brought closer together and dissimilar features are pushed apart.
Unlike fine-tuning on downstream tasks, unsupervised change detection using well pre-trained features relies on thresholding approaches and only considers bi-temporal images.
Discriminative models used in self-supervised change detection include pre-defined tasks and contrastive methods.
In \cite{leenstra2021self}, Leenstra et al. used predefined tasks for feature representation learning and trained a discriminative model to extract features from bi-temporal images for change detection.
An early attempt to use contrastive methods in change detection is \cite{chen2021self}, where the authors proposed contrastive learning for change detection in multi-view remote sensing images (including multi-temporal and multi-sensor images).
In \cite{chen2022self}, they further proposed a pixel-wise contrastive approach to distil the features to alleviate seasonal effects in bi-temporal change detection. 
One challenge these works face in using contrastive approaches to unsupervised pretraining is that they push apart samples that should belong to the same class, making it harder for the classifier to later categorize them correctly or create accurate decision boundaries.

\subsection{Change Detection in RS Image Time-series Using ConvLSTM}
Change detection is often associated with sequential data, making it necessary to evaluate temporal dynamics.
The computer vision community has addressed the modelling of temporal relationships among features using recurrent neural networks, which have proven effective for a wide range of applications such as object tracking and action recognition.
Long short-term memory networks (LSTM) are particularly effective for such problems, as they mitigate the vanishing gradient problem when dealing with long-term dependencies.
The combination of recurrent neural networks and deep learning architectures has also been used for time-series tasks, aiming to produce more useful feature representations by extracting both spatial and temporal information during the learning process.
Recent RS image time-series change detection tasks have extensively integrated LSTM techniques.
In \cite{mou2018learning}, an LSTM is integrated into a CNN to consider both spatial and temporal features in an end-to-end framework.
Sefrin et al. \cite{sefrin2020deep} proposed combining FCN and LSTM to study land-cover changes using Sentinel-2 images.
For high-resolution image change detection, Sun et al. \cite{sun2020unet} proposed using atrous Unet-ConvLSTM to better model multiscale spatial information.
For unsupervised approaches, Saha et al. \cite{saha2020change} treated change detection as an anomaly detection problem, using an LSTM network to learn a representation of the image time-series.
In this method, they used a pretext task of reordering the image sequence.
However, the predefined task cannot resist the influence of seasonal noise, which leads to many pseudo-changes in the results.
Some researchers have shown that pseudo-labels can help solve this problem.
Kalinicheva et al. \cite{kalinicheva2020unsupervised} proposed a new framework that combines a graph model and pseudo-labels, using a gated recurrent unit (GRU) AE-based model to associate the changes of consecutive images with different spatial objects.
Yang et al. \cite{yang2022utrnet} proposed an unsupervised time-distance-guided convolutional recurrent neural network (UTRnet) for change detection in irregularly collected images, using a weighted pre-change detection to obtain reliable training samples.

\begin{figure*}[pt]
	\centering
	\includegraphics[width=7.0 in]{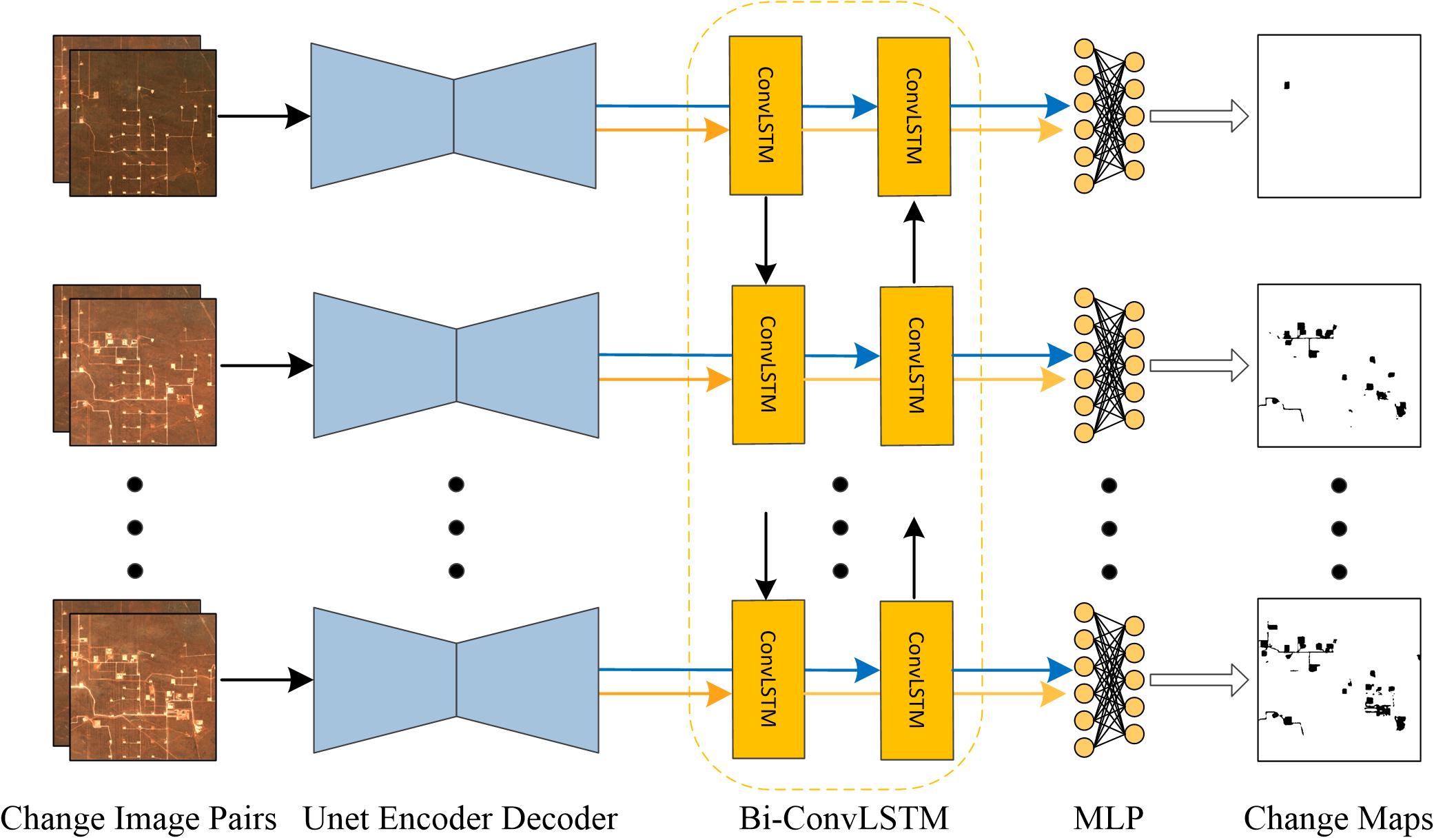}
	\caption{Overview of the proposed network architecture for RS image time-series change detection. The proposed network is based on the Unet and Bi-ConvLSTM. The change pairs are given as input to the Unet and then the output features are given as input to the Bi-ConvLSTM for temporal relationship modelling. At the end of the network, the output features of MLP are used as the inputs to the supervised contrastive loss, contrastive random walk loss and cross-entropy loss in the feature learning and finetuning stages.}
	\label{fig1}
\end{figure*}

\section{Methodology}
In this section, we present the proposed two-stage RS image time-series change detection framework. It includes a feature tracking-based pseudo label generation module and a self-training change detection module that follows the training setting of supervised contrastive learning. 
We first get the pixel-wise feature representation of each image in the image sequence using the pre-trained model \cite{chen2022self} and then get the pseudo change maps using the thresholding approach. Then, the feature tracking approach is used to get the final pseudo change labels based on the feature representation and the threshold-based change map.
Afterwards, the pseudo change labels are used to learn the representation of change maps using a supervised contrastive loss and the contrastive random walk loss. Finally, a fully connected layer is fine-tuned on learned change map representation using cross-entropy loss with a weighted supervised contrastive loss for final generating the change maps .  
In the following subsections, we will describe the network architecture of the proposed framework, the supervised contrastive loss, the contrastive random walk loss and the feature tracking-based pseudo label generation.

\subsection{Network Architecture}
The proposed approach uses an Unet-ConvLSTM network architecture, which consists of two components: ResUnet and Bi-ConvLSTM.
For the Unet, we adopt a similar architecture as the FC-Siam-conc \cite{daudt2018fully}.
It consists of two encoders, one bridge, one decoder, and skip connections between the downsampling and upsampling paths.
The decoder part has three blocks, each of which consists of a convolution layer (Conv), batch normalization (BN), ReLU, and upsampling.
A $1 \times 1$ Conv is used after the last block to reconstruct the learned representations.
We changed the padding type of all blocks to "same" padding.
The parameters and channel size of each unit are presented in Table \ref{table1}.
Each convolution unit ($[\cdot]$) includes a convolutional layer, a BN layer, and a ReLU activation layer.
Each residual block (ResBlk) in the encoding path has two residual units, each of which consists of two convolution units and an identity mapping.
\begin{table}[pt]
	\centering
	\caption{Structure of the proposed network.}
	\label{table1}
	\renewcommand\tabcolsep{7.0pt}
	\renewcommand\arraystretch{1.2}
	\begin{tabular}{cccc}
		\hline
		& Encoder 1 \& 2 & Decoder &  \\ \hline
		\begin{tabular}[c]{@{}c@{}}Conv1\\ Maxpool \end{tabular} & \begin{tabular}[c]{@{}c@{}}{[}3$\times$3, 32{]}, stride 2\\ 3$\times$3, stride 2 \end{tabular} &
		\begin{tabular}[c]{@{}c@{}}Cat. ResBlk2\\{[}3$\times$3, 128{]} \\upsampling 2 \end{tabular} &
		\begin{tabular}[c]{@{}c@{}}DecBlk1\\ stride 1\end{tabular} \\ \hline
		
		\begin{tabular}[c]{@{}c@{}}ResBlk1\\ stried 1\end{tabular} & $\left[\begin{array}{l} 3 \times 3,~32 \\ 3 \times 3,~32 \end{array}\right] \times 2$ &
		\begin{tabular}[c]{@{}c@{}}Cat. ResBlk1\\{[}3$\times$3, 128{]} \\upsampling 2 \end{tabular} &
		\begin{tabular}[c]{@{}c@{}}DecBlk2\\ stride 1\end{tabular} \\ \hline
		
		\begin{tabular}[c]{@{}c@{}}ResBlk2\\ stride 2\end{tabular} & $\left[\begin{array}{l} 3 \times 3,64 \\ 3 \times 3,64 \end{array}\right] \times 2$ &
		\begin{tabular}[c]{@{}c@{}}Cat. Conv1\\{[}3$\times$3, 128{]} \\upsampling 2 \end{tabular} &
		\begin{tabular}[c]{@{}c@{}}DecBlk3\\ stride 1\end{tabular} \\ \hline
		
		\begin{tabular}[c]{@{}c@{}}ResBlk3\\ stride 2\end{tabular} & $\left[\begin{array}{l} 3 \times 3,128 \\ 3 \times 3,128 \end{array}\right] \times 2$ & 
		\begin{tabular}[c]{@{}c@{}}{[}3$\times$3, 128{]} \\Bi-ConvLSTM \end{tabular} &
		\begin{tabular}[c]{@{}c@{}}LSTM Blk\end{tabular} \\ \hline
		\begin{tabular}[c]{@{}c@{}}Bridge\\ stride 1\end{tabular} & \begin{tabular}[c]{@{}c@{}}{[}3$\times$3, 256{]}\\ upsampling 2 \end{tabular}&
		\begin{tabular}[c]{@{}c@{}}{[}1$\times$1, 8/2{]} \end{tabular}&
		\begin{tabular}[c]{@{}c@{}}MLP\end{tabular}\\ \hline
		
	\end{tabular}
\end{table}
The output features of time-series change pairs are given in input the Bi-ConvLSTM layer.
Differently from the standard LSTM, ConvLSTM uses convolution operations in the input-to-state and state-to-state transitions to improve the modeling of the spatial correlation among sequence images.
It consists of an input gate $i_t$, an output gate $o_t$, a forget gate $f_t$, and a memory cell $\mathcal{C}_t$.
The input, output and forget gates act as controlling gates to access, update, and clear memory cell.
ConvLSTM can be formulated as follows (for convenience we remove the subscript and subscript from the parameters):
$$
\begin{array}{l}
	i_t=\sigma\left(\mathbf{W}_{x i} * \mathcal{X}_t+\mathbf{W}_{h i} * \mathcal{H}_{t-1}+\mathbf{W}_{c i} * \mathcal{C}_{t-1}+b_i\right) \\
	f_t=\sigma\left(\mathbf{W}_{x f} * \mathcal{X}_t+\mathbf{W}_{h f} * \mathcal{H}_{t-1}+\mathbf{W}_{c f} * \mathcal{C}_{t-1}+b_f\right) \\
	\mathcal{C}_t=f_t \circ \mathcal{C}_{t-1}+i_t \tanh \left(\mathbf{W}_{x c} * \mathcal{X}_t+\mathbf{W}_{h c} * \mathcal{H}_{t-1}+b_c\right) \\
	o_t=\sigma\left(\mathbf{W}_{x o} * \mathcal{X}_t+\mathbf{W}_{h o} * \mathcal{H}_{t-1}+\mathbf{W}_{c o} \circ \mathcal{C}_t+b_c\right) \\
	\mathcal{H}_t=o_t \circ \tanh \left(\mathcal{C}_t\right)
\end{array}
$$
where $*$ and $o$ denote the convolution and Hadamard functions, respectively.
$\mathcal{X}_t$ is the input tensor, $\mathcal{H}_t$ is the hidden state tensor, 
and, $\mathcal{W}_{x*}$ and $\mathcal{W}_{h*}$ are 2D convolution kernels corresponding to the input and hidden state, respectively, and $b_i$, $b_f$, $b_o$ and $b_c$ are the bias terms.

In this study, we employ Bi-ConvLSTM \cite{graves2005bidirectional,song2018pyramid} to encode the features of time-series change pairs.
It was proposed to use both past and future information to model sequential data.
Bi-ConvLSTM uses two ConvLSTMs to process the input data in both forward and backward directions, and then makes a decision for the current input by taking into account the data dependencies in both directions.
It has been shown that analyzing both forward and backward temporal perspectives improves predictive performance.
Each of forward and backward ConvLSTM can be considered as a standard one, with two sets of parameters for backward and forward states.
The output of Bi-ConvLSTM is calculated as follows:
\begin{equation}
	\mathbf{Y}_t=\tanh \left(\mathbf{W}_y^{\overrightarrow{\mathcal{H}}} * \overrightarrow{\mathcal{H}}_t+\mathbf{W}_y^{\overleftarrow{\mathcal{H}}} \overleftarrow{\mathcal{H}}_t+b\right)
\end{equation}
where $\overrightarrow{\mathcal{H}}_t$ and $\overleftarrow{\mathcal{H}}_t$ denote the hidden state tensors for forward and backward states, respectively, $b$ is the bias term, and $\mathbf{Y}_t$ indicates the final output considering bidirectional spatio-temporal information.
The hyperbolic tangent (tanh) is used to combine the output of both forward and backward states in a non-linear manner. 
After the last layer of Bi-ConvLSTM, an MLP block is used to reconstruct output features at the feature learning stage and predict the binary change maps at the finetuning stage.


\subsection{Loss Function}
During training, each image $I_i (i > 0)$ is used to construct a change pair anchored at the initial image ($I_{i=0}$).
The proposed network then predicts the change features of each change pair and the final change map, capturing the temporal changes related to the first image rather than the cumulated changes of the image sequence.
The training process uses a teacher-student paradigm and the exponential moving average (EMA) algorithm \cite{tarvainen2017mean}.
The input of the student network are the original time-series image pairs, while the teacher network uses the same time-series image pairs with color jitter.

According to supervised contrastive learning, the training process consists of the feature learning and finetuning stages.
In the feature learning phase, we use supervised contrastive loss with the contrastive random walk loss. The loss can be written as:
\begin{equation}
	\mathcal{L}_{\text {feat }}= \mathcal{L}_{sc} + \lambda_1 \mathcal{L}_{crw}
\end{equation}

In the finetuning stage, we use the cross-entropy loss with the weighted supervised contrastive loss.
The total loss is calculated as:
\begin{equation}
	\mathcal{L}_{\text {map }}=\mathcal{L}_{ce} + \lambda_2 \mathcal{L}_{sc}
\end{equation}
The $\mathcal{L}_{ce}$ is the cross-entropy loss, the $\mathcal{L}_{sc}$ is the supervised contrastive loss, and $\mathcal{L}_{crw}$ is the contrastive random walk loss.
The hyper-parameters $\lambda_1$ and $\lambda_2$ are used to tune the losses, and values of $\lambda_1 = 0.1$ and $\lambda_2 = 0.0005$ generally performed well in our experiments. 
In the following, we provide details on the supervised contrastive loss and the contrastive random walk loss.

\subsubsection{Contrastive loss}
The proposed approach uses a supervised contrastive loss \cite{khosla2020supervised} to differentiate representations between changed and unchanged pixels in time-series change pairs.
This loss is calculated by sampling over the pixel features in the constructed time-series pairs.
The pixel feature pairs at the same location in the output of the teacher and student networks are called positive pairs, while pixel features from different locations are called negative pairs.
Given a positive feature pair $(v_1^{i}, v_2^{i})$ and a pixel feature $v_2^j$ taken from another location, the contrastive loss can be formulated as $\mathcal{L}_{\text {contrast}}$:
\begin{equation}\label{eq3}
\mathcal{L}_{\text {contrast}}=-\underset{S}{\mathbb{E}}{\left[\log \frac {e^{sim(v_1^i, v_2^i)}}{{\sum_{j=1}^{N}} e ^{sim(v_1^i, v_2^j)}}\right]}
\end{equation}
where $sim$ is a similarity function (i.e., cosine similarity), $(v_1^i,v_2^i)$ is the normalized latent representation of pixel $i$, $(v_1^i,v_2^j|j\ge i)$ is the normalized latent representation of negative pair and $S=\{s_1^1, s_2^1,s_2^2, \cdots, s_2^{N-1}\}$ is a set that contains $N-1$ negative samples and one positive sample.
One limitation of self-supervised contrastive learning is that, since the class labels of the inputs are ignored, samples from the same class may end up being treated as negative pairs, which can affect the training performance.
To avoid this limitation and enable the contrastive loss to learn in a supervised fashion, Khosla et al. \cite{khosla2020supervised} extended the approach to account for input labels.
Following the original supervised contrastive learning method, we randomly sample $N$ pixel features in each change pair from the teacher-student network, generating two data views 
$\{(x_i, y_i)\}_{i= 1}^{2N}$, where $i \in I = [2N]$ is the index of an arbitrary sample.
Given $P = \{P_{i,j}|y_i = y_j, (x_i, y_i), (x_j, y_j)\}$, we perform supervised contrastive learning with sampled pixel features:
\begin{equation}
	\begin{aligned}
		\mathcal{L}_i^{\sup }= \sum_{i \in I} \frac{-1}{|P(i)|} \cdot \sum_{p \in P(i)} \log \frac{e^{\operatorname{sim}\left(x_i, y_p\right) / \tau}}{\sum_{b \in B(i)} e^{\operatorname{sim}\left(x_i, y_b\right) / \tau}}
	\end{aligned}
\end{equation}
where $B(i)$ means the set of indices excluding $i$, i.e., $B(i) = I \backslash {i}$; $P(i) = \{P_{i,j} | j \in B(i)\}$ is the positive set distinct from sample $i$ and $|\cdot|$ stands for cardinality.
In this case, the labels are binary pseudo labels.
The use of the supervised contrastive objective function improve the exploitation of the binary change information with respect to only use the cross-entropy loss.

\subsubsection{Contrastive Random Walk Loss}
Image time-series change detection is often treated as a simple extension of bi-temporal change detection in time.
However, the incorporation of temporal information to mitigate seasonal noise poses a significant challenge, because the change depicted at position $(x, y)$ in the frame $t$ might not have any relation to what we find at the same location $x, y$ in frame $t+k$.
To overcome this limitation, the contrastive random walk method leverages pathfinding on a space-time graph and associates features across space and time. 
This method establishes nodes shared by neighboring frames, thereby formulating correspondence.
By converting image time-series into palindromes, the walk step can be put through a contrastive learning problem. Moreover, the walker's destination offers guidance, allowing for the integration of whole chains of intermediate comparisons.

This works builds upon the contrastive random walk framework by \textit{et al.} \cite{jabri2020space}, where an image time-series is treated as a direct graph, composed of pixel feature vectors forming the nodes and weighted edges connecting neighboring frames.
Given an input image time-series with $k$ frames, we choose $N$ feature vectors $q_t$ within a small patch from the frame $t$, which serve as vertices of a graph.
The graph connects all feature vectors within the small patch in temporally adjacent frames. 
A random walker then steps through the graph, moving forward in time from frames $1, 2, \cdots, k$, and then backward in time from $k-1, k-2, \cdots, 1$, with transition probabilities determined by the similarity of learned representations.
Pairwise similarities are converted into non-negative affinities by applying a softmax function (with temperature $\tau$) over edges departing from each node.
This process generates the stochastic affinity matrix for the graph at each timestep.
\begin{equation}
	A_{t, t+1}(i,j) =\frac{exp(sim(q_t^i,q_{t+1}^j)/\tau)}{\sum_{l=1}^{N}exp(sim(q_t^i,q_{t+1}^j)/\tau)}
\end{equation}
for a pair of frames $t$ and $t+1$, where $q_t \in R^{N\times d}$ is the matrix of $d-$dimensional embedding vectors, $\tau$ is a small constant, and the softmax is performed along each row.
The local affinity between patches of two video frames, $q_t$ and $q_{t+1}$, is captured by this process.
The affinity matrix for the entire graph, which relates all nodes in the video as a Markov chain,  can be considered as a composition of local affinity matrices.
The established spatio-temporal connectivity in the graph and propose a walking strategy for a random walker that can perform tracking by contrasting similarity of neighboring nodes. 
Let $X_t$ to denote the state of the walker at time $t$, with transition probabilities $A_{t, t+1}(i,j) = P(X_{t+1}=j|X_t=i)$, where $P(X_t = i)$ is the probability of being at node $i$ at time $t$. With this view, we can formulate long-range correspondence as walking multiple steps along the graph:
\begin{equation}
	\overline{A}_{t, t+k} = \prod_{i=0}^{k-1} A_{t+i, t+i+1} = P(X_{t+k}|X_t)
\end{equation}
The likelihood of cycle consistency is maximized by training the model to achieve the event of the walker returning to its starting point. 
\begin{equation}
	L_{CRW} = - tr(log(\overline{A}_{t, t+k}\overline{A}_{t+k,t}))
\end{equation}
where $\overline{A}_{t, t+k}$ are the transition probabilities from frame $t$ to $t+k$: $\overline{A}_{t, t+k} = \prod_{i=t}^{t+k-1} A_{i, i+1}$.

\subsection{Pseudo Labels Extraction}
The pseudo change maps can be made less noisy by propagating the threshold-based pseudo labels to each change pair using the label propagation algorithm \cite{araslanov2021dense}.
This algorithm considers both spatial and temporal neighbours, using a queue of the $N_T$ most correlated change pairs for temporal neighbours, and a spatial neighbourhood of the query node for spatial neighbours.
The labels of target nodes are determined by computing the matrix of transitions between target nodes and source nodes, considering only top$-k$ transitions, and multiplying it by the labels of the source nodes.
For every feature embedding in a frame, we compute its cosine similarity with the features in the queue, and select the $N_k=10$ feature embeddings with the highest similarity.
We then use these embeddings to compute a weighted sum of the label predictions at these locations, which is added to the label context.
This process is repeated for all change pairs in the image sequence, updating the label and embedding contexts using the $N_T$ most correlated change pairs of each change pair.

In detail, we obtain the embedding for change pair $t$, denoted as $h_t \in R^{h, w, n}$, where $h$, $w$ are the spatial dimensions and $n$ is the dimension of the feature embedding, using a pre-trained network.
The embedding context defined as $\mathcal{E}_t=\left\{h_0, h_{t-N_T+1}, \ldots, h_{t-1}\right\}$ maintains embedding of each change pair.
Similarly, we define the label context as $M_t =\left\{m_{0}, m_{t-N_{T}+1}, \cdots, m_{t-1}\right\}$ obtained by using threholding method on pre-trained features.
We also define the predicted labels as $L_t = \left\{l_{0}, l_{t-N_{T}+1}, \cdots, l_{t-1}\right\}$.
We compute the cosine similarity of embedding $h_t(i,j)$ with all embeddings in queue $\mathcal{E}$, restricted to a spatial-temporal neighbourhood of size $N_T \times P \times P$, centred on location $(i, j)$, which we denote as $N_p(i,j)$.
We then obtain the nearest-neighbour set $N_p^{(k)}(i,j)$ by selecting $N_k$ embedding locations from $N_P(i,j)$ with the highest cosine similarity and compute the local weight as follows:
\begin{equation}
s_{i, j}^{t}\left(t^*, l, n\right)=
		\frac{\exp \left(-d_{i, j}^{t}\left(t^*, l, n\right) / \tau\right)}{\sum_{\left(t^{\prime}, l^{\prime}, n^{\prime}\right)} \exp \left(-d_{i, j}^{t}\left(t^{\prime}, l^{\prime}, n^{\prime}\right) / \tau\right)}
\end{equation}
The three coordinates $(t, l, n)$ for the temporal (first) and the spatial (second and third) dimensions specify the neighbour locations in $N_p(i,j)$.
$d_{i,j}^{t}(t^*,l,n)$ is the cosine similarity between embeddings $h_{t}(i,j)$ and $h_{t^*}(l,n)$ from $N_p^{(k)}(i,j)$; $(t^*,l,n)$ and $(t^{\prime}, l^{\prime}, n^{\prime}) \in N_p^{(k)}(i,j)$ , and $\tau$ is the temperature hyperparameter set to 0.005.
We compute the label $l_t$ as a weighted sum of the label predictions in $N_p^{(k)}(i,j)$ as
\begin{equation}
	l^t(i,j)=\sum_{\left(t^*, l, n\right) \in N_P^{(k)}(i, j)} s_{i, j}^{t}\left(t^*, l, n\right) m^{t^*}(l, n)
\end{equation}
where $m^{t^*}(l,n)$ comes from the label context $M$.
We repeat this process for the remaining change pairs in the image sequence.

\section{Experiments}
In this section, we first describe the datasets used in our experiments and then introduce the related experiment setting on the network training and the feature-tracking-based pseudo-label generation. Finally, we present the results of the proposed approach and the comparison methods. We also present an ablation study of each component of the proposed approach.
\subsection{Description of Datasets}
We conducted experiments on two multi-spectral datasets, one from the Sentinel-2 satellite constellation and the other from the Landsat-8 satellite.

\subsubsection{Sentinel-2 dataset}
The Multi-temporal Urban Development (MUDS) dataset \cite{van2021multi} was designed to monitor urbanization by tracking changes in building construction from 2017 to 2020. It is an open-source dataset that includes native Planet 4-meter resolution imagery and Sentinel-2 multi-spectral images with irregular observation intervals across six continents. However, the original Sentinel-2 imagery often contains clouds and missing values. To improve the utility of this dataset, we selected only 74 locations with a minimum of 12 clean images and resized each image to 512 x 512 pixels. Of these 74 locations, we labelled the significant change pair of 30 scenes for future evaluation, with all change pairs referenced to the first image. We used only four bands in this work, all of them with a spatial resolution of 10 meters. Due to the unsupervised nature of this dataset, we only considered three types of changes: built-up, bare land, and water.

\subsubsection{Landsat-8 dataset}
The UTRnet dataset \cite{yang2022utrnet} was specifically designed for validation of the UTRnet model. The dataset consists of the satellite image time-series collected by Landsat-8 from 2013 to 2021, with a spatial resolution of 30 meters. Six spectral bands covering the visible to the shortwave infrared region are used, including blue, green, red, near-infrared, and two shortwave infrared bands. The dataset includes nine typical scenes located in different cities in China, each with a different land cover type. For each scene, ten cloud-free Landsat-8 images were selected to cover different seasons. The image size for each scene is 400 x 400 pixels. The ground truth includes three classes: changed pixels, unchanged pixels, and unlabeled pixels. The changed and unchanged pixels are labelled using Google Earth images. In this study, unlabeled pixels are treated as unchanged pixels to validate the influence of seasonal noise. Due to the temporal limitations of high-resolution image labelling, the labels only include the longest interval pairs. The change maps include city expansion, water change, and soil change.

\subsection{Experiment Settings}
\subsubsection*{Evaluation Metrics}
In order to evaluate the effectiveness of different methods in binary change detection, this paper employs five evaluation metrics: precision (Pre), recall (Rec), overall accuracy (OA), F1 score (F1), and Cohen's kappa score (Kap). 

\subsubsection*{Implementation Details}
In the process of generating pseudo labels, we first derived pseudo labels of each change pair using a thresholding approach on pre-trained features. Then propagate the threshold-based labels to each change pair using the feature tracking approach. In the setting of feature tracking parameters, the spatial neighbours $P$ are set to 10, the temporal neighbours $N_T$ are set to 3 most correlated change pairs, and the value of $N_k$ is set to 10.
In the self-training algorithm, the proposed approach uses a two-layer Bi-ConvLSTM. 
We choose the Adam optimizer with an initial learning rate of $3e^{-4}$ at feature learning stage, which is decreased using step scheduling without restarts. The batch size is set to 2 and the model is trained for 200 epochs. 
For the finetuning, we use SGD optimizer with a learning rate of 0.01, a mini-batch size of 10 and a number of epochs equal to 50.
To evaluate the proposed approach, it is compared with the state-of-the-art method UTRnet in fitting and unseen scenarios. UTRnet is an improved LSTM-based self-training approach that uses CVA to generate pseudo labels. Unlike the proposed approach, UTRnet is not designed to generalize to unseen scenarios and requires fitting a separate model for each scene. 
In the evaluation, we choose the fitting evaluation on the Landsat-8 dataset due to the lack of training data while choosing the inference evaluation on the Sentinel-2 dataset.
For the fitting scenarios of the Landsat-8 dataset, we chose scene 3, scene 5 and scene 7 as evaluation sets. For the unseen scenarios of the Sentinel-2 dataset, we chose scene $T1286\_2921\_13$, scene $T1736\_3318\_13$ and scene $T6730\_3430\_13$ as the evaluation set.
In addition to the comparison with UTRnet, this paper also conducts extensive ablation experiments on the labelled 30 scenes of the Sentinel-2 dataset to evaluate the impact of different components of the proposed approach and different pseudo-label generation algorithms. In particular, the proposed approach is compared with versions that do not use the contrastive random walk loss or only use the cross-entropy loss.  
It should be noted that only the change pair with the most significant change is labelled as ground truth for evaluating the performance of different approaches.

\begin{figure*}[pb]
	\centering
	\includegraphics[width=7.0 in]{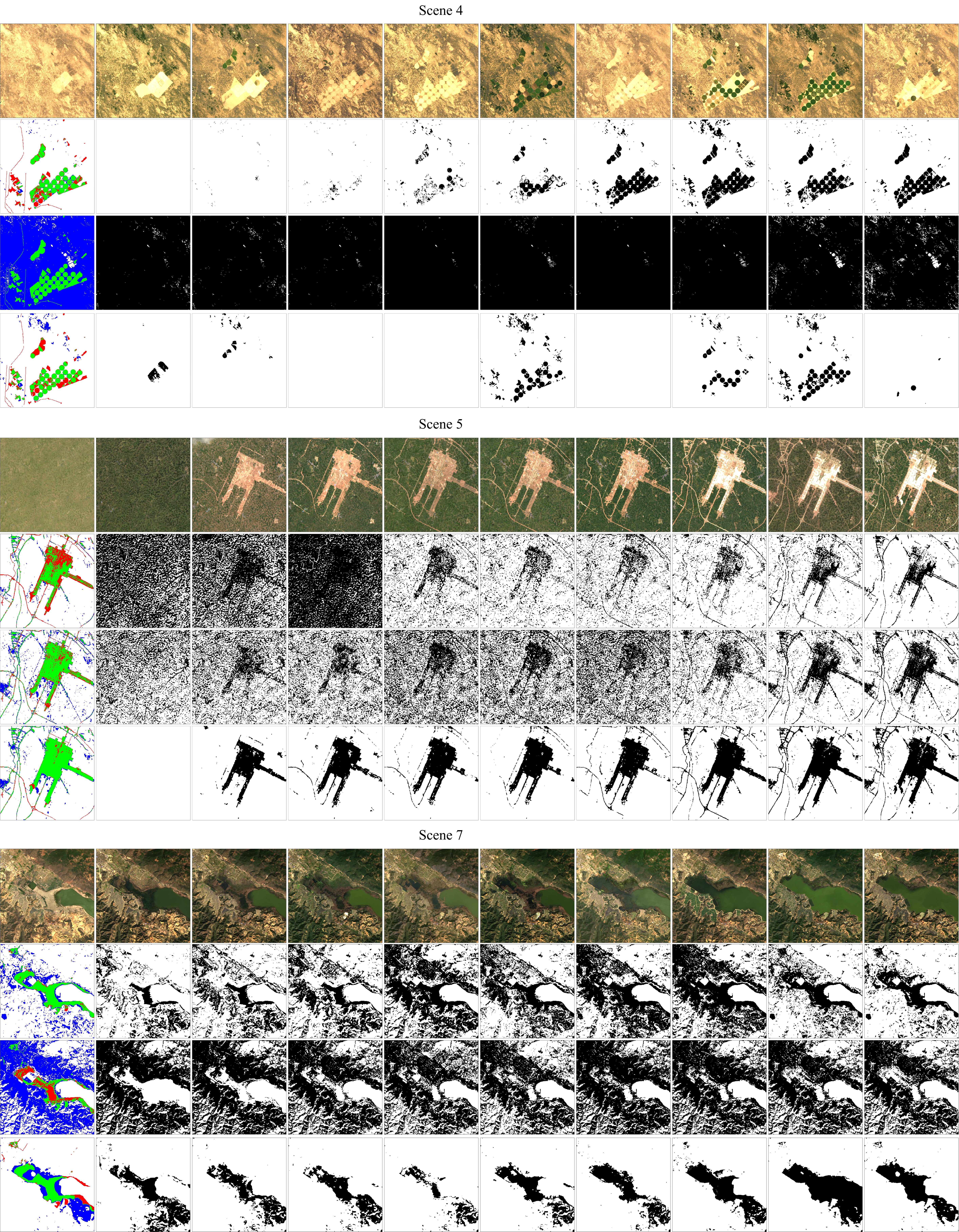}
	\caption{Examples of change detection results on three scenes for the Landsat-8 dataset. Row 1: image time-series; Row 2: change maps of one-scene fitting obtained by UTRnet; Row 3: change maps of all-scene fitting obtained by UTRnet; Row 4: change maps of all-scene fitting obtained by the proposed approach. Col. 1 of Row 2, 3, 4 in each scene is the most significant change map versus the ground truth (Green: TP, White: TN, Blue: FN, Red: FP).}
	\label{fig2}
\end{figure*}

\subsection{Experimental Results on Landsat-8 image time-series}
In this study, the effectiveness of the proposed approach is evaluated using the Landsat-8 dataset.
The performance of the proposed approach is compared with the state-of-the-art approach UTRnet, which has been validated by fitting on each scene in the dataset.
In order to evaluate the generalizability of the proposed approach, results are provided for fitting on all scenes, while UTRnet results are provided for both fitting on each scene and fitting on all scenes.
Quantitative evaluation is performed using the most significant change in the time-series of images, due to the challenges of differentiating changes in continuous change scenarios.
The results of the proposed approach and UTRnet are presented in Table \ref{table2}.
In the one-scene fitting setting, UTRnet achieves an OA of 89.66\% and a Cohen's kappa score of 0.56, underperforming the results obtained from the pseudo labels.
However, for the all-scene fitting setting, UTRnet fails to differentiate changed and unchanged pixels, achieving an OA of 48.50\% and a Cohen's kappa score of 0.09.
In contrast, the proposed approach achieves significantly better results than UTRnet in both settings, with an OA of 91.30\% and a Cohen's kappa score of 0.64.
Comparing the results of the pseudo labels and the models, we can see that self-training approaches further improve the results of the corresponding pseudo labels.
It is worth noting that the pseudo labels acquired by weight-CVA are even better than those of the proposed approach in the significant change pairs, but they result in a worse model performance due to the lack of enough change pairs for training.

Besides the quantitative analysis, we also provide a visual comparison of the results obtained from the proposed approach and the UTRnet method.
We present the results of UTRnet obtained by fitting on each scene, and the results of UTRnet obtained by fitting on all scenes as well as the results of the proposed approach.
We also present the significant change maps in each first column of the change maps in each scene in Fig \ref{fig2}, where true positives, true negatives, false negatives, and false positives are colored in green, white, blue, and red, respectively.
From the visual comparison of the most significant change map, we can see that the change map obtained by UTRnet using all-scene fitting is noisy and contains a high number of false alarms.
In contrast, the change maps obtained by the other two settings are more accurate and have less noise.
Additionally, the proposed approach is able to successfully detect most of the changed pixels and suppress the effects of seasonal changes.
When comparing the change map time-series, we can see that the change maps obtained by UTRnet (one-scene fitting) have more false alarms that are affected by historical changes.
In contrast, the change maps obtained by the proposed approach are robust to seasonal changes and only focus on real changes happened at each time.
While the one-scene fitting UTRnet still achieves good results on all test scenes, the all-scene fitting UTRnet can still perform well in a few scenes with less seasonal noise, but its results are heavily influenced by the imbalanced training samples.

\begin{table}[pt]
	\centering
	\caption{Quantitative evaluations of different approaches applied to the fitting test set on the Landsat-8 dataset.}
	\label{table2}
	\renewcommand\tabcolsep{5.0pt}
	\centering
	\begin{tabular}{ccccccc}
		\hline
		& Method & Pre(\%) & Rec(\%) & OA(\%) & F1 & Kap \\ \hline
		& Weighted CVA & 61.89 & 67.80 & 91.02 & 0.647 & 0.596 \\
		& Feature Tracking & 53.11 & 76.01 & 88.07 & 0.625 & 0.557 \\
		& UTRnet (One-Scene) & 60.86 & 62.57 & 89.66 & 0.617 & 0.557 \\
		& UTRnet (All-Scene)  & 17.50 & 77.19 & 48.50 & 0.285 & 0.087 \\	
		& Proposed.(All-Scene) & 64.89 & 74.76 & 91.30 & 0.695 & 0.644 \\ \hline
	\end{tabular}
\end{table}

\begin{table}[pt]
	\centering
	\caption{Quantitative evaluations of different approaches applied to the inference test set on the Sentinel-2 dataset.}
	\label{table3}
	\renewcommand\tabcolsep{5.5pt}
	\centering
	\begin{tabular}{ccccccc}
		\hline
		& Method & Pre(\%) & Rec(\%) & OA(\%) & F1 & Kap \\ \hline
		& Weighted CVA & 91.02 & 51.99 & 92.54 & 0.662 & 0.623 \\
		& Feature Tracking & 55.18 & 92.97 & 88.02 & 0.693 & 0.624 \\
		& UTRnet (One-Scene) & 85.25 & 25.17 & 88.89 & 0.388 & 0.347 \\
		& UTRnet (Inference)  & 46.64 & 64.07 & 84.67 & 0.540 & 0.451 \\	
		& Proposed. (Inference) & 76.12 & 84.46 & 93.90 & 0.801 & 0.765 \\ \hline
	\end{tabular}
\end{table}

\begin{figure*}[pt]
	\centering
	\includegraphics[width=6.5 in]{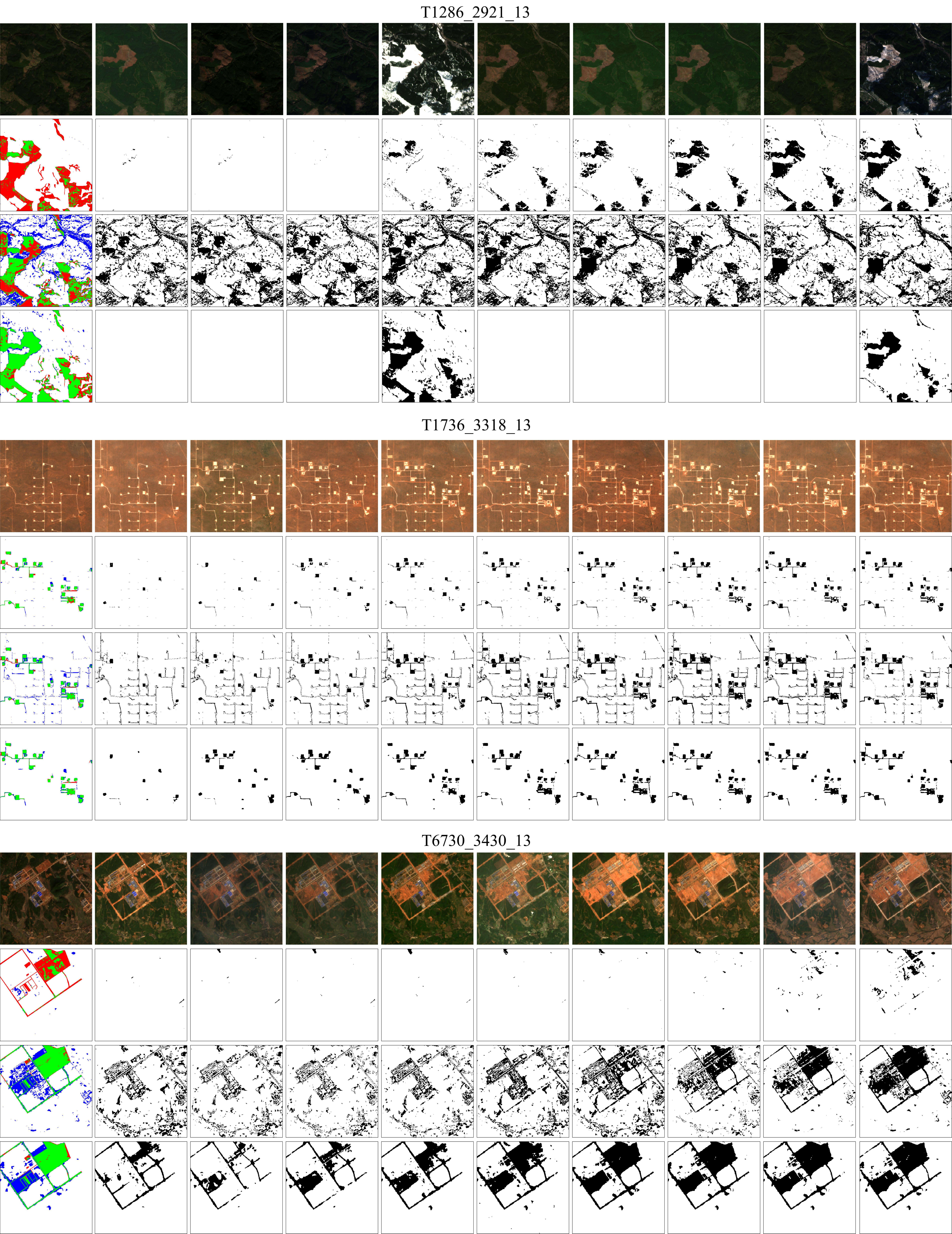}
	\caption{Examples of change detection results on three scenes for the Sentinel-2 dataset. Row 1: image time-series; Row 2: change maps of one-scene fitting obtained by UTRnet; Row 3: change maps on infer setting obtained by UTRnet; Row 4: change maps on infer setting obtained by the proposed approach. Col. 1 of Row 2, 3, 4 in each scene is the most significant change map versus the ground truth (Green: TP, White: TN, Blue: FN, Red: FP).}
	\label{fig3}
\end{figure*}

\subsection{Experimental Results on the Sentinel-2 image time-series}
The Sentinel-2 dataset is characterized by its diversity of land-cover scenes and a larger number of training samples.
In contrast to the results obtained on the Landsat-8 dataset, we present the results of different pseudo labels, one-scene fitting UTRnet, and inference on unseen scenarios based on models trained on all training samples.
Similar to the evaluation on the Landsat-8 dataset, we only consider the most significant change map in each scene to assess its quantitative performance (Table \ref{table3}).
As one can observe, the one-scene fitting UTRnet achieves worse results than those obtained on the Landsat-8 dataset, with an OA of 88.89\% and a Cohen's kappa score of 0.35.
The possible reason is that the Sentinel-2 dataset contains more seasonal changes such as snow.
However, its performance is improved when inferred to unseen scenarios.
Nevertheless, it still shows significant improvements compared to the all-scene fitting setting on the Landsat-8 dataset, which is largely due to the increased number and diversity of training samples.
On the other hand, the inference results obtained from the proposed approach are significantly better than those obtained by UTRnet in both the one-scene fitting and inference on unseen scenarios settings.
Across all five performance metrics, the proposed approach achieves the best performance in most cases, except for precision, achieving an OA of 93.9\% and a Cohen's kappa score of 0.77. 
This indicates that the proposed approach not only outperforms the state-of-the-art approach on trained samples, but also on unseen samples.
In the proposed approach the improvement is more pronounced when using a larger and more diverse set of training samples.
Similarly to the experiments on the Landset-8 dataset, UTRnet underperforms the results of the pseudo labels.
In this case, two pseudo-label generation approaches achieve comparable accuracy.

In addition to the quantitative analysis, we also provide a visual comparison of the most significant change map and the change map time-series in each scene.
Fig. \ref{fig3} shows a comparison of all methods on the Sentinel-2 test set.
The true positive, true negative, false negative, and false positive pixels of the significant change map are colored green, white, blue, and red, respectively.
We first analyze the performance of the most significant change map in each scene.
As shown in the figure (first column of change maps in each scene), the proposed approach successfully detects most changed pixels and suppresses seasonal changed areas while the results of UTRnet contain more false alarms and missing detections.
For the change map time-series, one can see that the results obtained by one-scene fitting UTRnet contain many missed detection in particular related to the cultivated errors in the image sequence.
As for the inference results, UTRnet fails to suppress most seasonal changes and presents more false alarms, but gets big improvements in noise reduction compared with its performance on the Landset-8 dataset.
This issue is well addressed in the proposed approach, where abrupt changes and continuous changes are both well detected.

\begin{figure*}[pt]
	\centering
	\includegraphics[width=7.0 in]{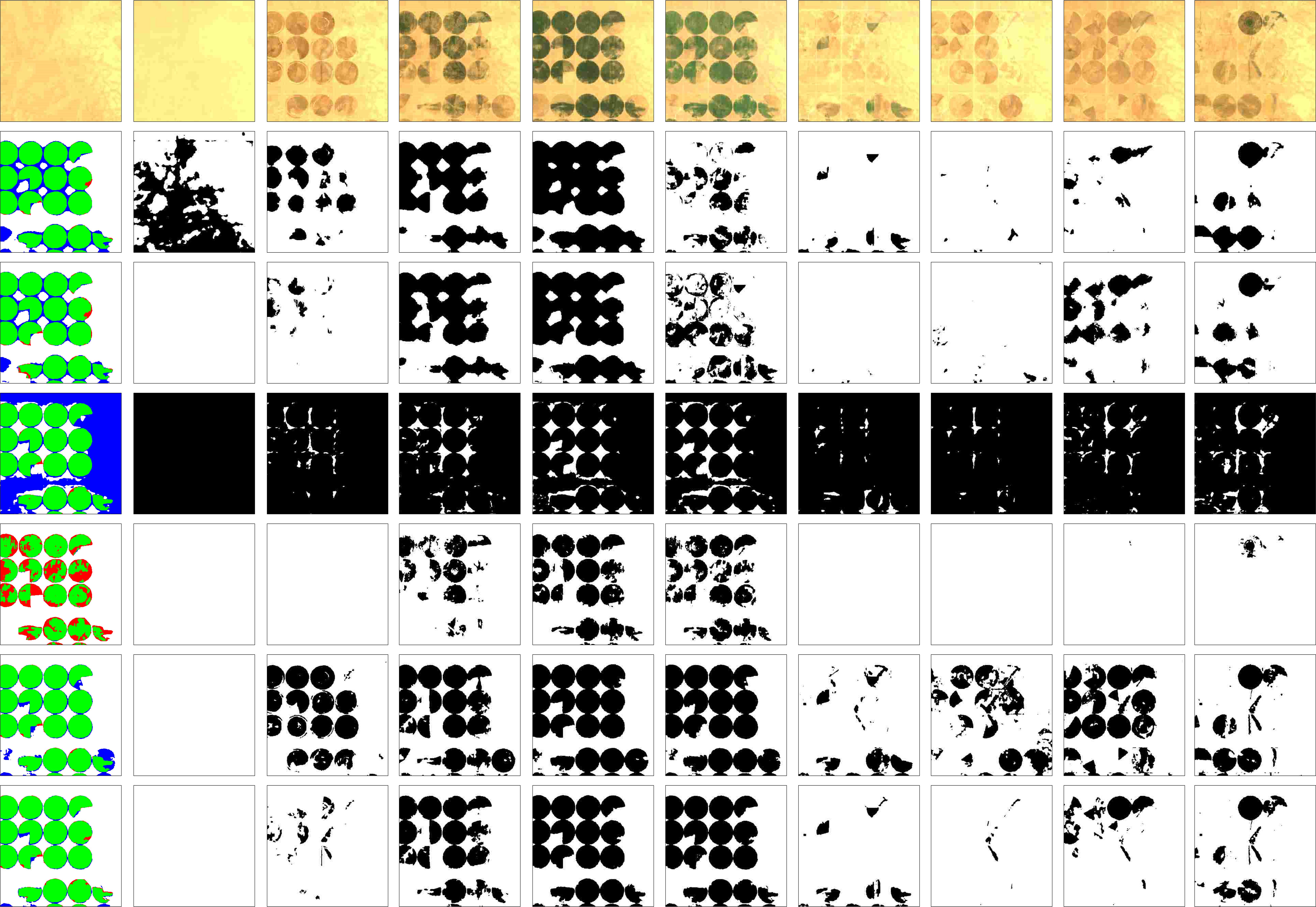}
	\caption{Examples of change detection results on the Sentinel-2 ablation test set. Row 1: image time-series; Row 2: pseudo labels obtained by thresholding approach; Row 3: pseudo labels obtained by feature tracking; Row 4: change maps obtained by the proposed approach only using cross-entropy loss; Row 5: change maps obtained by the proposed approach only using contrastive loss; Row 6: change maps obtained by the proposed approach trained on threshold-based pseudo labels; Row 7: change maps obtained by the proposed approach trained on feature tracking-based pseudo labels. Col. 1 of Row 2-7 is the most significant change map versus the ground truth (Green: TP, White: TN, Blue: FN, Red: FP).}
	\label{fig4}
\end{figure*}

\subsection{Discussion}
In this section, we conduct extensive ablation studies on the proposed approach to analyze the contribution of different components.
To better understand the proposed approach, we choose the scene $T4780\_3377\_13$ with significant vegetation changes over time for visualization. However, the quantitative evaluation was implemented on the selected 30 scenes of the Sentinel-2 dataset as the ablation test set.

\subsubsection{Pseudo labels}
Many unsupervised change detection approaches employ a thresholding approach for change detection.
However, determining a reasonable threshold is often a challenging task.
To demonstrate the effectiveness of the proposed pseudo-label generation approach, we present the pseudo-labels obtained by thresholding and feature tracking methods, individually.
Then, we train the proposed approach using these two sets of pseudo-labels.
Finally, we evaluate the performance of the trained models on the ablation test set.
Fig. \ref{fig4} shows the details of the pseudo-labels and the results obtained by the trained models.
As one can see, the thresholding approach produces more false alarms in the pseudo change maps with the shorter time interval change pair.
In contrast, the feature tracking approach can mitigate the effect of this type of seasonal changes while maintaining the most significant changes in the change map time-series.
Similarly, the model trained on threshold-based labels produces more missing detections due to this type of noise, while the model trained on feature tracking-based labels significantly reduces the false alarms.
Table \ref{table4} presents all five metrics on the ablation test set for the two trained models.
Among these results, the model trained on feature tracking-based labels provides the best result in almost all metrics, including the highest overall accuracy of 94.41\% and the highest kappa coefficient of 0.599.
Compared to the threshold method-based labels, the OA and Kappa on the feature tracking-based labels are further improved by about 2\% and 0.05.
This indicates that the feature tracking-based pseudo-label generation can more accurately detect reliable changes in RS image time-series and thus benefit the self-training.

\subsubsection{supervised contrastive loss and contrastive random walk loss}
To verify the effectiveness of the contrastive and contrastive random walk loss, we set up experiments with training on the proposed pseudo-labels.
Specifically, we trained models using both supervised contrastive and contrastive random walk losses, and only using supervised contrastive loss, as well as only using cross-entropy loss, respectively.
The same ablation test set defined before is used for evaluation.
Fig. \ref{fig4} and Table \ref{table4} compare the results obtained using three different models.
Results show that the supervised contrastive loss and contrastive random walk loss achieve significant improvements in noise reduction and maintain the consistency of changes in the time-series.
The only use of the contrastive loss achieves an OA of 93.67\% and a kappa of 0.553, which are slightly lower than the values obtained by using two loss functions.
In addition, the use of both loss functions increases by about 6\% and 0.16 on OA and Kappa, respectively, with respect to the only use of the cross-entropy loss.
This demonstrates that the joint use of contrastive loss and contrastive random walk loss can further improve the performance of the self-training paradigm.

\begin{table}[pt]
	\centering
	\caption{Quantitative evaluations of different approaches applied to the Sentinel-2 test set in the ablation study.}
	\label{table4}
	\renewcommand\tabcolsep{4.5pt}
	\centering
	\begin{tabular}{ccccccc}
		\hline
		& Method & Pre(\%) & Rec(\%) & OA(\%) & F1 & Kap \\ \hline
		& threshold-based & 38.30 & 19.18 & 90.78 & 0.256 & 0.212 \\
		& feature-tracking & 40.27 & 17.13 & 91.07 & 0.240 & 0.201 \\
		& only cross-entropy & 37.94 & 74.17 & 87.86 & 0.502 & 0.441 \\
		& only contrastive & 63.48 & 54.68 & 93.67 & 0.588 & 0.553 \\
		& self-training thres. & 53.72 & 66.22 & 92.51 & 0.593 & 0.552 \\
		& self-training feature. & 69.53 & 57.47 & 94.41 & 0.629 & 0.599 \\	 \hline
	\end{tabular}
\end{table}

\section{Conclusion}
In this work we have proposed a new framework for detecting changes in RS image time-series without any manually annotated training data.
Our framework jointly uses an architecture based on Unet and ConvLSTM and adopts a self-training algorithm.
We first extract pseudo labels using the feature-tracking method and then further improve the results by training a model from scratch.
Feature-tracking-based pseudo label generation in RS image time-series detect the significant changes more accurately while alleviate the presence of seasonal changes.
The self-training combines the use of supervised contrastive loss, contrastive random walk loss and cross-entropy loss following the two-stage setting of supervised contrastive learning.
This mitigates the effects of the noise in pseudo labels and keep the consistency of change map time-series.
Our experiments on two different datasets demonstrate the effectiveness of our approach compared to state-of-the-art methods.
It is worth noting that the proposed approach can also generalize well to unseen scenarios.
Although our method is demonstrated in the context of multi-spectral images, it can be applied to other sensors, such as synthetic aperture radar and RGB images.
In future work, we plan to extend our method to detect different types of changes using prior information from multi-spectral images.


\ifCLASSOPTIONcaptionsoff
  \newpage
\fi

\bibliographystyle{IEEEtran}
\bibliography{mylib}

\end{document}